Соломія Бук, Андрій Ровенчак
Україна, Львів

# ОНЛАЙН-КОНКОРДАНС РОМАНУ ІВАНА ФРАНКА «ПЕРЕХРЕСНІ СТЕЖКИ»

**Вступ**

Серед різноманіття лексикографічних праць виокремлюють особливий тип словника — конкорданс, — що подає до кожного реєстрового слова (чи словоформи) усі або вибіркові контексти його вживання, демонструючи таким чином принаймні мінімальне його лексичне оточення.

Зважаючи на це, конкорданс є відправною точною для визначення конкретної семантики лексеми. Як зазначав Ж. Вандрієс, "в усіх випадках значення слова визначається контекстом. Слово ми ставимо в оточення, що виявляє його значення, кожен раз і на даний момент. Не що інше, а саме контекст, всупереч різноманітності значень даного слова, надає йому "особливого" значення; не що інше, а саме контекст, очищує його від колишніх значень, нагромаджених пам'яттю, і створює йому "актуальне" значення" [11: 251]. М. Кочерган додає, що контекст стосовно слова виконує такі найголовніші функції: є засобом відбору та актуалізації потрібного значення; модифікує смисл у межах одного значення, уточнює його, розширюючи або обмежуючи клас денотатів; служить засобом синкретизації значень полісемічного слова; формує оказіональне значення; виступає засобом десемантизації [11: 251–252]. Отже, справедливим видається твердження: "Сьогодні виглядає на те, що нові словники укладатимуть на основі конкордансів" [22: 169].

"В основі к[онкордансу] завжди лежить певний закінчений текст чи закрита однорідна група таких текстів (твір, ряд творів, усі твори одного автора)" [6: 267]. Цим пояснюють факт, що більшість відомих конкордансів різних мов побудовано на матеріалі творів конкретного письменника [12; 17; 24]. Враховуючи ті переваги в опрацюванні текстової інформації, які має комп'ютер, дослідники звичайно і укладають, і подають інформацію конкордансів в електронній формі.

Через складність укладання (до появи комп'ютерів) перші конкорданси створювали вручну і лише для особливо важливих текстів, насамперед для Біблії, Корану, а згодом і для творів деяких видатних авторів, як-от У. Шекспіра. При цьому контексти, як правило, супроводжували коментарями. Відколи стала можливою автоматична обробка текстів, саме Біблія залишається найпоширенішим джерелом конкордансів різними мовами [14; 15; 19]. На сучасному етапі добре опрацьованою і представленою в мережі Інтернет є творчість У. Шекспіра [17] і Дж. Джойса [24]. Чи не єдиним російським автором, до творів якого вдалося виявити онлайн-конкорданс, є Ф. Достоєвський [12].

Перші окремі елементи конкордансу в українській лексикографії можна помітити у словнику В. Ващенка "Епітети поетичної мови Т. Г. Шевченка" у розділах "Епітетне пояснення слів у контекстах Шевченкових поезій" та "Функціональний обсяг епітетів у контекстах", де, відповідно, до кожного іменника вказано епітети, які з ним поєднував Т. Шевченко і навпаки: до кожного епітета подано іменники, з якими він сполучається у творах "Кобзаря". [4: 72–79]. Весь матеріал опрацьовано вручну, сам словник доступний тільки в паперовому вигляді.

Пізніше укладено чотиритомний конкорданс поетичних творів Тараса Шевченка, який "точно відмічає, де в "Кобзарі", і в яких текстуальних обставинах появляється кожна вжита Шевченком словоформа, як українська, так і російська" [10, Т. 1: XIX]. У ньому зафіксовано 18 401 лексичну одиницю і подано 83 731 випадки їхнього вживання. Опрацювання матеріалу здійснено з використанням комп'ютерних програм, проте сам словник доступний тільки в паперовому форматі.

Автори зазначають, що їхня праця — "перша конкорданція в українському літературознавстві", таким чином дещо звужують сферу застосування своєї праці, не розповсюджуючи інформацію, яку можна почерпнути зі словника, на лінгвістику. Проте, слід зазначити, що це подія у цілій філологічній науці, адже конкорданс (в ідеалі побудований не лише до поетичних, а до всіх творів письменника), який подає синтагматику кожного слова, — без перебільшення, неоціненне джерело знань також і для мовознавців, а саме для виявлення особливостей ідіостилю автора, мовної та концептуальної картини світу, відображеної в його творчості тощо.

Поряд із задумом комплексного квантитативного опису творчості І. Франка та розробкою теорії створення його корпусу текстів [1], окремої уваги заслуговує й укладання конкордансу Франкових текстів. Це непросте й трудомістке завдання, яке ускладнюється відсутністю академічного видання, а також обмеженою кількістю текстів, доступних в електронному вигляді. Тому на даний момент видається можливим поетапне опрацювання окремих творів, конкорданси яких згодом можна об'єднати.

Першим об'єктом нашого дослідження став роман "Перехресні стежки", який автори досліджували раніше з метою вивчення статистичних параметрів [16] і укладання частотного словника [3].

**Методика укладання конкордансу роману**

Для створення конкордансу ми використовували результати роботи над укладанням частотного словника "Перехресних стежок" [3]. З метою технічного опрацювання тексту було розроблено оригінальний пакет програм. Вихідними матеріалами були:
- частотний список слів у початковій формі, із вказівкою на частину мови кожного, у такому форматі:

| абсолютна частота | лема | частина мови |
|---|---|---|
| 2851 | І | AC |
| 360 | Й | AC |
| 2471 | ВІН | P |
| 2248 | НЕ | AP |
| 1331 | В | AR |
| 832 | У | AR |
| 1728 | Я | P |
| 1508 | НА | AR |
| 1222 | З | AR |
| 176 | ІЗ | AR |
| 62 | ЗІ | AR |
| 40 | ЗО | AR |
| 1360 | ЩО(спол.) | AC |
| 1303 | БУТИ | V |
| 1220 | ТОЙ | P |
| 1146 | СЕЙ | P |
| 1073 | ДО | AR |
| 1065 | А(спол.) | AC |
| 903 | ВОНА | P |

Непослідовність з погляду частот розташування деяких форм пов'язана з тим, що у частотному словнику вони об'єднані в одну словникову статтю (І / Й, З / ІЗ / ЗІ / ЗО і т. д.). Дієслова позначено літерою V (*verb*), іменники — N (*noun*), прикметники — J (*adjective*), прислівники — D (*adverb*), числівники — M (*numeral*), займенники — P (*pronoun*). Службові частини мови кодуються дволітерною комбінацією (AC (*conjunction*) — сполучник, AR (*preposition*) — прийменник, AP (*particle*) — частка, AI (*interjection*) — вигук, AA (*article*) — артикль).

- алфавітний список словоформ із вказаними лемами. Для зручності роботи цей список було поділено на частини відповідно до початкової літери. Третя і п'ята колонки містять абсолютні частоти словоформ і лем відповідно, що дає змогу контролювати правильність лематизації.

| | | | | |
|---|---|---|---|---|
| Б | Б | 89 | Б | 89 |
| БА | БА | 4 | БА | 4 |
| БАБА | БАБА | 5 | БАБА | 9 |
| БАБИ | БАБИ | 2 | БАБА | |
| БАБИНЕЦЬ | БАБИНЕЦЬ | 3 | БАБИНЦІ(село) | 7 |
| БАБИНЦІВ | БАБИНЦІВ | 1 | БАБИНЦІ(село) | |
| БАБИНЦЯХ | БАБИНЦЯХ | 3 | БАБИНЦІ(село) | |
| БАБІЙ | БАБІЙ | 4 | БАБІЙ(прізв.) | 4 |
| БАБСЬКИМ | БАБСЬКИМ | 1 | БАБСЬКИЙ | 1 |
| БАБУ | БАБУ | 2 | БАБА | |
| … | | | | |
| БАРАНОВІ# | БАРАНОВІ(прикм.) | 1 | БАРАНІВ(прикм.) | |
| БАРАНОВІ | БАРАНОВІ(ім.) | 5 | БАРАН(прізв.) | |

На підставі цих двох списків за допомогою програми `lemmas.cgi` створювалися файли `*.lemma` відповідності словоформа–лема у форматі

| | | |
|---|---|---|
| А## | А(виг.) | AI |
| А | А(спол.) | AC |
| А# | А(част.) | AP |
| АБИ | АБИ | AC |
| АБІХТ | АБІХТ(ім'я) | N |
| АБНЕҐАЦІЇ | АБНЕҐАЦІЯ | N |
| АБО | АБО(спол.) | AC |
| АБО# | АБО(част.) | AP |
| АБСОЛЮТИСТИЧНОЇ | АБСОЛЮТИСТИЧНИЙ | J |
| АБСОЛЮТНА | АБСОЛЮТНИЙ | J |
| АБСОЛЮТНО | АБСОЛЮТНО | D |
| АБСОЛЮТНОЇ | АБСОЛЮТНИЙ | J |
| АВАНГАРДУ | АВАНГАРД | N |
| АВАНС | АВАНС | N |
| АВАНСУ | АВАНС | N |
| АВАНСУВАВ | АВАНСУВАТИ | V |
| АВАНСУВАЛИ | АВАНСУВАТИ | V |
| АВАНТЮРИ | АВАНТЮРА | N |
| АВАНТЮРУ | АВАНТЮРА | N |
| АВЖЕЖ | АВЖЕЖ | AP |

…

Перша колонка відповідає словоформі, як вона функціонує в тексті; друга — відповідній лемі, за потреби з коментарями в дужках; у третій колонці містяться граматичні показники (на даному етапі це лише вказівка на частину мови, у перспективі передбачено доповнити цей опис іншими граматичними категоріями). Знак "#" використано у тексті для розрізнення омографів.

Для маркування тексту було розроблено програму `marking.cgi`, яка на підставі об'єднаного файлу `.lemma` створювала розмічений текст за таким форматом:

`словоформа<ідентифікатор частини мови|ЛЕМА(за потреби з уточненнями)>`

Приклад:
-- А<AP|А(част.)>, пан<N|ПАН> меценас<N|МЕЦЕНАС>! Ґратулюю<V|ҐРАТУЛЮВАТИ>, ґратулюю<V|ҐРАТУЛЮВАТИ>! Може<V|МОГТИ> тішитися<V|ТІШИТИСЯ> наше<P|НАШ> місто<N|МІСТО>, що<AC|ЩО(спол.)> дістало<V|ДІСТАТИ> такого<P|ТАКИЙ> блискучого<J|БЛИСКУЧИЙ> оборонця<N|ОБОРОНЕЦЬ>. О<AI|О(виг.)>, такої<P|ТАКИЙ> оборони<N|ОБОРОНА> наш<P|НАШ> трибунал<N|ТРИБУНАЛ> давно<D|ДАВНО> не<AP|НЕ> чув<V|ЧУТИ>!
Се<P|СЕЙ> було<V|БУТИ> на<AR|НА> вулиці<N|ВУЛИЦЯ>, перед<AR|ПЕРЕД> будинком<N|БУДИНОК> карного<J|КАРНИЙ> суду<N|СУД>, в<AR|В> однім<M|ОДИН> із<AR|ІЗ> більших<J|БІЛЬШИЙ> провінціональних<J|ПРОВІНЦІОНАЛЬНИЙ> міст<N|МІСТО>. Власне<AP|ВЛАСНЕ(част.)> вибила<V|ВИБИТИ> перша<M|ПЕРШИЙ>, карна<J|КАРНИЙ> розправа<N|РОЗПРАВА> скінчилася<V|СКІНЧИТИСЯ>, і<AC|І> з<AR|З> суду<N|СУД> виходили<V|ВИХОДИТИ> купами<N|КУПА> свідки<N|СВІДОК> -- селяни<N|СЕЛЯНИ>, жиди<N|ЖИДИ>, якісь<P|ЯКИЙСЬ> ремісники<N|РЕМІСНИК>, поліційні<J|ПОЛІЦІЙНИЙ> стражники<N|СТРАЖНИК>. Адвокат<N|АДВОКАТ> д-р<N|Д-Р> Євгеній<N|ЄВГЕНІЙ(ім'я)> Рафалович<N|РАФАЛОВИЧ(прізв.)> вийшов<V|ВИЙТИ> також<D|ТАКОЖ>, вирвавшися<V|ВИРВАТИСЯ> з-поміж<AR|З-ПОМІЖ> своїх<P|СВІЙ> клієнтів<N|КЛІЄНТ>, цілої<J|ЦІЛИЙ> купи<N|КУПА> селян<N|СЕЛЯНИ>
...

Інформація, подана у кутових дужках, називається "тег" (tag). Зазначену схему маркування за потреби можна легко звести до вимог TEI (Text Encoding Initiative) — Міжнародного стандарту в цій галузі [23].

Іншомовні слова лематизовано за загальними правилами, при цьому переклад (наведений у виданні [20, 173–459] як підсторінкова примітка) в електронному варіанті тексту подано у фігурних дужках. Цей текст не включено до конкордансу, оскільки його авторство належить редакторам 50-томника, а не І. Франкові (у виданні 1900 р. [13] ці переклади відсутні), відповідно, він не виводиться при зверненні до веб-сторінки конкордансу:

...
Він<P|ВІН> був<V|БУТИ> дуже<D|ДУЖЕ> задоволений<J|ЗАДОВОЛЕНИЙ>, але<AC|АЛЕ>, держачися<V|ДЕРЖАТИСЯ> старого<J|СТАРИЙ> правила<N|ПРАВИЛО> "aequam<J|AEQUUS(лат.)> servare<V|SERVO(лат.)> mentem<N|MENS(лат.)>" {зберігати рівновагу духу (лат.)}, мав<V|МАТИ(дієсл.)> вид<N|ВИД> не<AP|НЕ> то<AP|ТО> байдужно-спокійний<J|БАЙДУЖНО-СПОКІЙНИЙ>,
...

Для коректного відображення діакритичних знаків у німецьких, польських, французьких словах використано спеціальні засоби мови HTML (ä для ä і под.) з однією різницею — знак ";" у цих послідовностях замінено з технічних міркувань (оскільки ";" виконує в тексті функцію розділового знака) на "^":
... остільки<D|ОСТІЛЬКИ> "урядові<J|УРЯДОВИЙ> шпички<N|ШПИЧКА>" (так<D|ТАК> перекладав<V|ПЕРЕКЛАДАТИ> Стальський<N|СТАЛЬСЬКИЙ(прізв.)> німецький<J|НІМЕЦЬКИЙ> термін<N|ТЕРМІН> Spitzen<N|SPITZE(нім.)> der<AA|DIE(нім.озн.арт.мн.)> Behö^rden<N|BEH\"ORDE(нім.)> {Найвище начальство (нім.)})...

За допомогою програм alph.cgi і alphclean.cgi створено файли *.html для кожної літери окремо, які містять алфавітний список слів (лем) з гіперпокликами на спеціальний скрипт concord.cgi, який ґенерує web-сторінку за запитом.
У лексикографії склалося дві основних форми подання контексту у конкордансах: 1) наведення однакової кількості слів до і після заданого слова [20] (як варіант — літер [12]); 2) наведення повного речення, у якому функціонує лексема [15; 18]. Як виняток, можна зустріти й наведення цілого вірша або сцени драматичного твору, у якому трапилось реєстрове слово [21]. Це питання безпосередньо пов'язане із визначенням обсягу розпізнавального контексту, необхідного і достатнього для ідентифікації значення лексеми, а також із локалізацією уточнювальних засобів і характеру значення, що уточнюється [5: 252].

Оптимальним з погляду співвідношення мінімального обсягу тексту та максимального висвітлення в ньому значення нам видається подання контексту пореченнєво із попереднім та наступним реченнями включно (у межах одного абзацу). Зважаючи на те, що багато речень у Франковій прозі є досить довгими (що незручно опрацьовувати), у паралельному вікні ми запропонували дещо спрощене формальне розв'язання цього питання (+/– 7 слів до і після лексеми у межах одного абзацу — т. зв. KWIC — Key Word In Context [22: 173]), що, на нашу думку, є зручним у користуванні.

Формат виводу інформації у конкордансі "Перехресних стежок" показано на прикладі кількох перших словоформ леми "Я":

КОНКОРДАНС ЛЕМИ **Я** (форма контексту — речення):

1. -- Що, не пізнають **мене** пан меценас? -- говорив він радісно і дуже голосно, немов бажав, щоб і прохожі чули його слова. -- А, не диво, не диво!
2. Ще й як бачились! Ану, прошу придивитися **мені** добре, прошу пригадати собі, га, га, га!..

…

6. Пан меценас ще тут незнайомі… **Я** тут офіціал при помічнім уряді, маю під собою регістратуру. О, я служу вже п'ятнадцять літ!
7. Я тут офіціал при помічнім уряді, маю під собою регістратуру. О, **я** служу вже п'ятнадцять літ!

КОНКОРДАНС ЛЕМИ **Я** (форма контексту — +/– 7 слів):

| # | лівий контекст | слово | правий контекст |
|---|---|---|---|
| 1 | -- Що, не пізнають | **мене** | пан меценас? – говорив він радісно і |
| 2 | Ще й як бачились! Ану, прошу придивитися | **мені** | добре, прошу пригадати собі, га, га, га!.. |
| 3 | Будьте ласкаві, допоможіть моїй пам'яті! Їй-богу, стидно | **мені,** | але ніяк не можу… |
| 4 | -- Ах! Ото з | **мене** | забудько! Пан Стальський, мій домашній інструктор у |
| 5 | -- Авжеж, авжеж! | **Я** | в суді. Пан меценас ще тут незнайомі… |
| 6 | в суді. Пан меценас ще тут незнайомі | **Я** | тут офіціал при помічнім уряді, маю під |
| 7 | помічнім уряді, маю під собою регістратуру. О, | **я** | служу вже п'ятнадцять літ! |
| 8 | -- Так. Власне тоді, як | **я** | пана меценаса вчив, мене з шестої класи |
| 9 | Власне тоді, як я пана меценаса вчив, | **мене** | з шестої класи відібрали до війська. Дурний |
| 10 | чоловік був. Було шануватися, зістати офіцером Ну, | **я** | там зразу троха шарпався Знаєте, у війську |
| 11 | Знаєте, у війську мусить бути субординація. Так | **я** | й став на фельфеблю. А вислуживши десять |
| 12 | став на фельфеблю. А вислуживши десять літ, | **я** | пішов і дістав місце канцеліста при суді |

Конкорданс виводить на екран відразу всі контексти функціонування заданого слова, почергово їх нумеруючи, так що користувач відразу має інформацію про абсолютну частоту вживання слова. Такий формат подання матеріалу для користувача є більш практичним порівняно з форматом, коли певну обмежену кількість контекстів заданого слова на екрані висвітлено посторінково (пор. [20]).

Експериментальна версія онлайн-конкордансу [2] міститься за адресою http://www.ktf.franko.lviv.ua/~andrij/science/Franko/concordance.html.

Вона на даний час передбачає виконання двох основних функцій: 1) пошук за словом (для цього слід зайти на відповідну літеру і клацнути на потрібному слові); 2) пошук за словоформою чи її частиною.

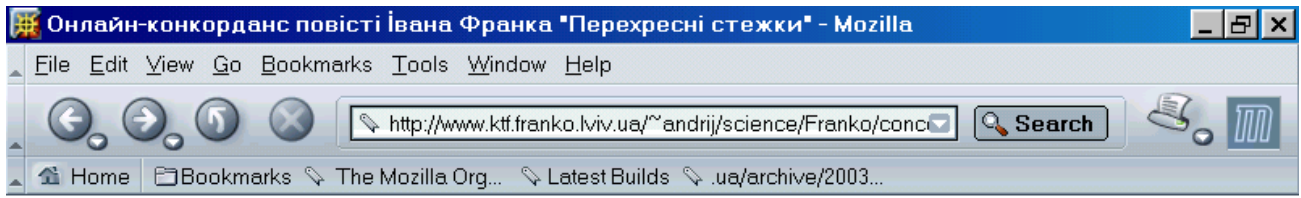

**Висновки**

Оскільки текст роману повністю анотовано за допомогою граматичних тегів, на цьому матеріалі можна виконувати різні дослідження, наприклад, пошук слів за частиною мови, що, зокрема, дозволяє виявити особливості їх функціонування у тексті І. Франка (див. теґи на позначення частин мови). Такий текст є також джерелом даних для лінгвостатистичного аналізу.

У перспективі передбачено виділення в романі фразеологізму як окремої одиниці тексту (що уможливить їх автоматичний пошук за будь-яким словом з його складу), здійснення синтаксичного маркування та ін.

Практичне значення конкордансу найяскравіше виявляється у можливості його використання для створення Словника мови Івана Франка, ідею якого розробив професор Львівського університету І. Ковалик [7; 8; 9]. Зважаючи на величезний обсяг творчості І. Франка в різних родах і жанрах (від поезії до перекладів та наукових статей з економіки), такий словник став би неоціненною енциклопедією і мови, і життя другої половини ХІХ — початку ХХ ст. Адже тільки на основі контексту можна визначити те значення, яке слово реалізує у тексті. Електронна форма словника, по-перше, дозволяє замінити трудомісткий етап ручного карткування слів і, по-друге, дає можливість автоматичного виявлення усіх синтагм зі словом за лічені секунди.

Конкорданс також є зручним інструментом для виявлення фрагментів концептуальної картини світу автора, а саме емотивного, ціннісного, просторового, соціального та ін. просторів тексту; особливостей індивідуальномовної картини світу і культурно-історичної, етнічної специфіки ментальності автора; може бути використаним для створення загальної теорії тексту.

**Література**

Резюме
У статті описано теоретичну методику та практичну реалізацію укладання конкордансу роману Івана Франка «Перехресні стежки». Запропоновано дві форми подання контексту вживання слів. Розроблено електронну версію цієї лексикографічної праці, яка доступна для загального користування в Інтернеті.

Ключові слова: конкорданс, контекст, лема, теґ

**Online-concordance of Ivan Franko's novel *Perekhresni stežky* (*The Cross-Paths*)**
Solomiya Buk, Andrij Rovenchak

Summary
In the article, theoretical principles and practical realization for the compilation of the concordance to Ivan Franko's novel *Perekhresni stežky* (*The Cross-Paths*) are described. Two forms for the context presentation are proposed. The electronic version of this lexicographic work is available online.

Key words: concordance, context, lemma, tag